\documentclass[letterpaper, 10 pt, conference]{ieeeconf}

\IEEEoverridecommandlockouts                              

\usepackage{graphics} 
\usepackage{epsfig} 
\usepackage{mathptmx} 
\usepackage{times} 
\usepackage{amsmath} 
\usepackage{amssymb}  
\usepackage{color}
\usepackage{hyperref}
\usepackage{cite}
\usepackage{float}
\usepackage{comment}
\usepackage{soul}
\usepackage{xcolor}
\usepackage{ulem}
\usepackage{algorithmic}
\usepackage[font=small,labelfont=bf]{caption}
\usepackage{subcaption}
\title{\LARGE \bf

Teleoperated Omni-directional Dual Arm Mobile Manipulation Robotic System with Shared Control for Retail Store
}

\author{Rolif Lima$^{*1}$, Somdeb Saha$^{1}$,  Nijil George$^{1}$, Vismay Vakharia$^{1}$, Shubham Parab$^{1}$,  \\Sahil Gaonkar$^{1}$, Vighnesh Vatsal$^{1}$ and Kaushik Das$^{1}$
\thanks{$^{1}$The authors are with TCS Research, Tata Consultancy Services Ltd., Bengaluru, Karnataka - 560066, India. *Corresponding Author: \tt\small rolif.lima@tcs.com}
}

\begin{document}

\maketitle
\thispagestyle{empty}
\pagestyle{empty}

\begin{abstract}
The swiftly expanding retail sector is increasingly adopting autonomous mobile robots empowered by artificial intelligence and machine learning algorithms to gain an edge in the competitive market. However, these autonomous robots encounter challenges in adapting to the dynamic nature of retail products, often struggling to operate autonomously in novel situations. In this study, we introduce an omni-directional dual-arm mobile robot specifically tailored for use in retail environments. Additionally, we propose a tele-operation method that enables shared control between the robot and a human operator. This approach utilizes a Virtual Reality (VR) motion capture system to capture the operator's commands, which are then transmitted to the robot located remotely in a retail setting. Furthermore, the robot is equipped with heterogeneous grippers on both manipulators, facilitating the handling of a wide range of items. We validate the efficacy of the proposed system through testing in a mockup of retail environment, demonstrating its ability to manipulate various commonly encountered retail items using both single and dual-arm coordinated manipulation techniques.

\end{abstract}

\section{INTRODUCTION}
The fast-paced growth of the retail industry, with stores expanding in size, increasing product ranges, and proliferating global presence is simultaneously affected by the shortage of labour in developing nations, competition from other retailers, and the rise of e-commerce. These conditions have led to the widespread adoption of automation technology in this industry to maximise the efficiency of operations and increasing profit margins by boosting sales and minimising the losses incurred. 

One such action taken to enhance efficiency comprises of use of mobile robots in the retail store to automate several monotonous and tedious operations which otherwise would require human labour. These systems have already been adopted for performing a wide range of tasks in the current retail store such as, "Whiz", a robot with vacuum cleaning ability is deployed in more than 2500 stores across Japan \cite{rindfleisch2022robots},  A shelves scanning robots by Walmart and Braincorp for inventory management, assistive robots such as "Pepper" and "Lowebot" for answering customer queries \cite{donepudi2020robots}. The need for such robotic systems has also driven competitions such as the "Amazon picking challenge" and "Future convenience store challenge", leading to vast contributions towards the development from the research community.

Some of the primary tasks involved in the retail store involve, anomaly correction in product placements (typically caused by customers), disposal of expired goods, and restocking shelves with items from the inventory storage. To perform such operations autonomously, the robot is required to have ability to navigate it's environment autonomously while avoiding collision with static and dynamic obstacles such as the aisles and customers in the store. Additionally, perception is required to be capable of detecting, recognizing and localizing a large variety of items and also be able to easily adapt to new items that are regularly introduced to the store, similarly the adaptive gripping mechanism along with a manipulator is essential to grasp the items efficiently without damaging them.

\begin{figure}[t]
\centering
\includegraphics[width=0.45\textwidth,trim={0cm 0.75cm 0 1cm},clip]{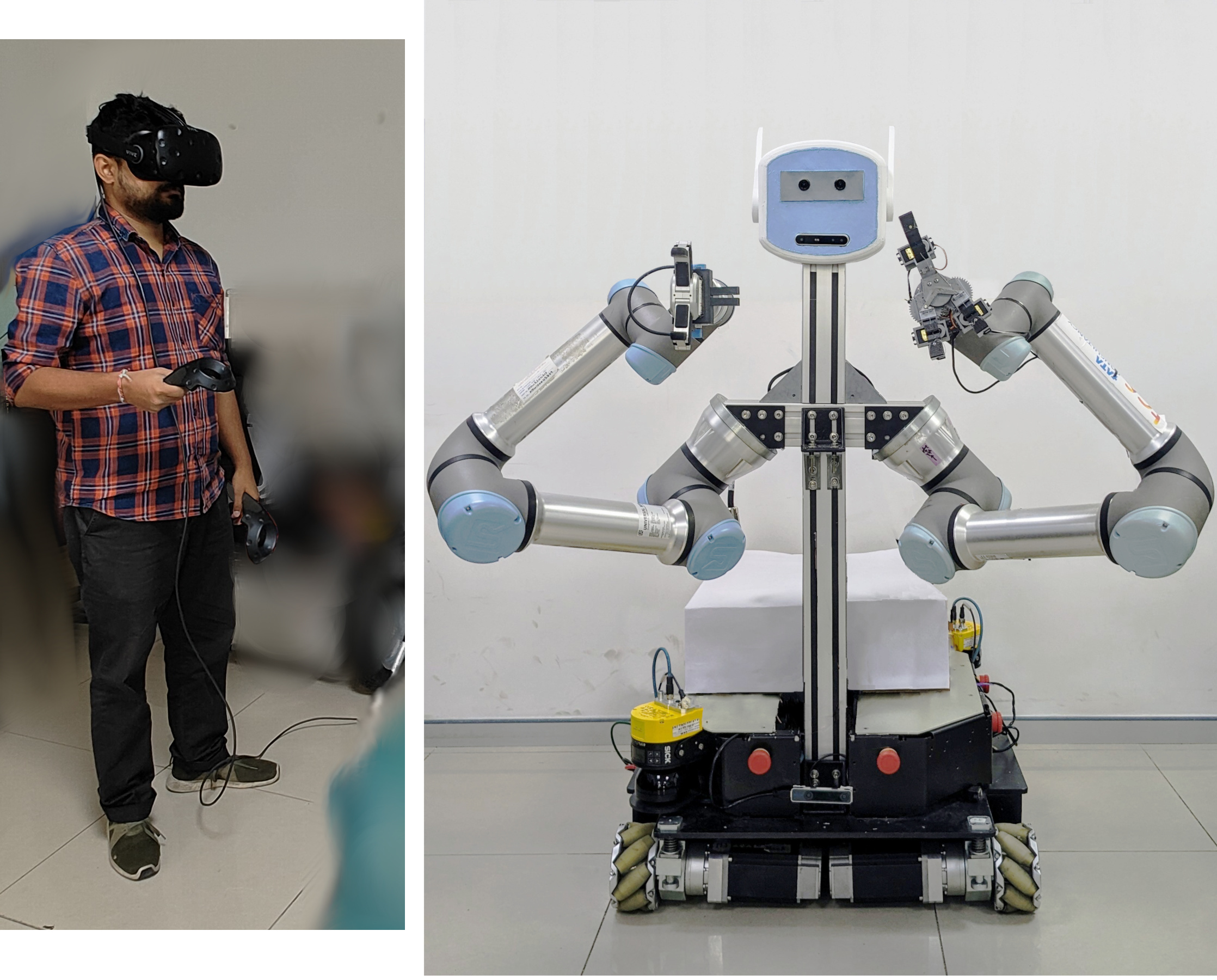}
\caption{(Left) Operator wearing head-mounted display and carrying two controllers. (Right) GriffinX: An in-house built omni-directional mobile robot with a dual-arm collaborative manipulators}
\label{fig:intro_fig}
\end{figure}

Numerous autonomous approaches have been proposed by different research groups, demonstrating capabilities in areas such as shelf restocking and anomaly correction. The work in \cite{mronga2024marlin} presents cloud integrated mobile-robot solution for intra logistic activities in a retail store, wherein, AI and robotics services such as digital representation of retail stores, data analysis, symbolic reasoning and process planning are obtained from the cloud service. The proposed solution lacks manipulation capabilities and assist the in-store employee in placing the items at appropriate shelfs using a pointer unit.

A dual-arm robot for intralogistics is presented in \cite{9212558}, wherein the main focus is on the identification of manoeuvres that are commonly adopted by a human worker in the retail store for picking items from densely packed shelves, further it uses the dual-arm manipulator equipped with two different grippers to replicate these motions to perform similar tasks. Similarly, a system with autonomous capabilities to localise and grasp the items from the shelf using a custom-designed universal vacuum gripper(UVG) and the ability to autonomously navigate while avoiding collision with humans is presented in a \cite{kai2024development,sakai2020mobile}. Similarly, a robotic system with a compact compliant pneumatic gripper for manipulating tightly packed items in a retail store with on-place straightening algorithm is proposed is \cite{garcia2020restock}. 
 
As can be seen from the literature robots with autonomous navigation capabilities utilizing several sensing mechanisms are capable of navigating retail store-like environments, however the manipulation capabilities demonstrated by the existing papers are limited to simplified scenarios tailored to specific conditions and cannot be generalized to all the intricacies that exit in a commercial retail store \cite{sengar2022challenges}. Thus in this work, we propose a robotic system for a retail store with tele-operation capabilities with shared control architecture.

Teleoperation of robots has been a widely studied subject due to it's ability to allow its users to manipulate remote environments via a robot while the operator is at a distant geographical location \cite{darvish2023teleoperation}.  However, with regards to the robots with tele-operation capabilities have been mostly restricted to customer interaction  such as in ~\cite{song2021teleoperated} to enhance the sales by means of shoping assistance, or to perform simple tasks using telepresence such as in  ~\cite{Kangaroo}, and not much works have been reported in the area of tele-operation with shared autonomy in retails store settings.

In this study, we present a dual-arm omni-directional mobile manipulation  robotic system called "GriffinX" aimed at retail store intra-logistic operations. Secondly we propose teleoperation with shared control feature aimed at enhancing the operator's effectiveness in executing manipulation tasks via teleoperation specific to retail environments, solely relying on visual feedback from the robot as shown in Fig. \ref{fig:intro_fig}. The impetus for this research stems from the dynamic nature of retail settings, where the introduction of new items or alterations in packaging occurs daily, posing challenges to robots relying solely on AI and ML algorithms for autonomous operation. Such situations may result in the failure of the robot to complete tasks due to unfamiliar scenarios not included in its training data. In these instances, the teleoperation functionality allows a remotely situated operator to intervene and successfully complete the task. Moreover, the corrective actions taken by the operator can serve as valuable feedback for refining the robot's capabilities, enabling it to autonomously handle similar encounters in the future.

\subsection{Contributions}
Main contributions of this work are as listed below
\begin{itemize}
    \item A omni-directional mobile manipulation system with dual arm manipulators each equipped with unique gripper, designed specifically for a retail scenario.
    \item A Tele-operation of the mobile manipulator with shared control framework for controlling the dual arm manipulators to perform single arm and coordinated dual arm manipulations.
    \item Integration of the proposed tele-operation algorithm with the developed dual arm mobile robot
\end{itemize}
  Further the proposed system is tested for it's efficacy in a mockup retail store by performing manipulation of several retail store products. 

Rest of the paper is organised a follows, detailed description of the proposed system architecture is presented in Sec. \ref{sec:sysArchi}, followed by description of teleoperation and shared control formulation in Sec. \ref{sec:teleopeAndSA}. Sec. \ref{sec:sysEval} presents the results corresponding to the evaluation of the system in the mockup retail scenario followed by conclusion and future works in Sec. \ref{sec:conclusion}.

\begin{figure*}[h!]
    \centering
	\includegraphics[width=0.9\linewidth]{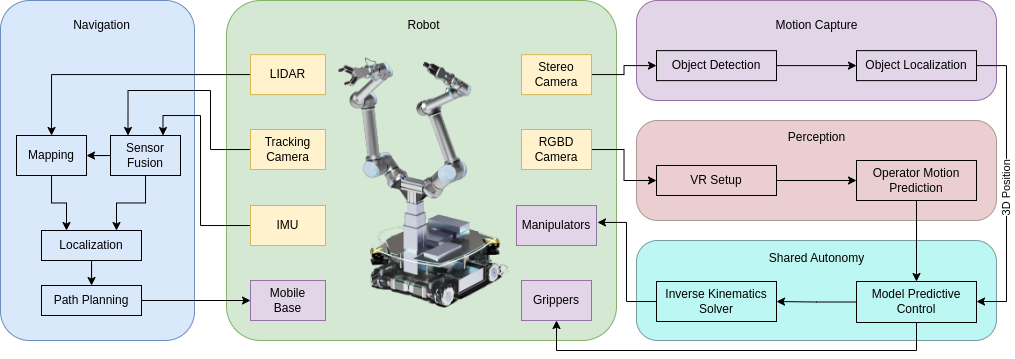}
	\caption{System architecture} 
	\label{fig:sysArchi}
\end{figure*}

\section{SYSTEM ARCHITECTURE}
\label{sec:sysArchi}
Detailed schematic of the system architecture is shown in Fig. \ref{fig:sysArchi} and detailed explanation of all the components is given in this section below.
\subsection{Motion Capture}
A motion capture setup is used to extract the position and orientation of the operator's arm movements. This can be achieved by using several techniques such as the use of an exoskeleton, a master system having a similar kinematic structure as the robot, or using extrinsic sensing such as RGB cameras, depth cameras, or  virtual reality (VR) setup~\cite{desmarais2021review,lee2022inertial,menolotto2020motion}.

In this study, we use a VR setup (HTC Vive) to capture the operator's arm position and orientation with respect to the head-mounted display using the handheld VR controllers, and the available switches to trigger the gripping action and drive the mobile platform.

\subsection{Mobile Robot with Dual Arms}
Retail store environments commonly have flat, smooth floors with narrow aisles. Considering these requirements, an omnidirectional platform is designed and fabricated in-house to be used as a mobile base for the robto. It provides agility, efficiency, and reliability in autonomous navigation within retail environments. The chassis is designed to house the associated electronic components, batteries, and sensors. The mecanum wheels coupled with high-torque BLDC motors are used to provide the omni-directional capability with precise motion control of the base platform. It allows to traverse forward, backwards, diagonally and in sideways directions as well as in-place rotation; thus providing exceptional agility and making it suitable for retail store applications. The mobile base is equipped with an NVIDIA Jetson Xavier TX2 as an onboard computer to perform critical control and navigation tasks by interfacing with a suite of sensors. For obstacle avoidance and navigation, LiDAR, Real-Sense depth camera D415, and T265 self-tracking cameras, IMU and wheel encoders are used. 



For manipulation in the retail store, two Universal Robots UR5e Cobots (collaborative robots)~\cite{sherwani2020collaborative}, each offering 6 degrees of freedom (DoF) are mounted on the mobile platform. The manipulators are mounted in a symmetrical configuration at 45 degrees relative to the base.


\subsection{Gripper}
 The end-effector is one of the most important parts of a manipulation system and its effectiveness determines the overall success rate of the entire system. In this work, we have incorporated two different types of grippers as end-effectors to the two manipulators to enable the robot to handle a wide variety of store items- a two-fingered rigid gripper and a novel three-fingered soft gripper. 

\subsubsection{OnRobot RG2 Gripper}
We are using an OnRobot RG2 two-fingered rigid gripper. Two-finger rigid grippers are easier to control and plan grasp for and is used for grasping a variety of standard objects.
\begin{figure}[t]
\centering
\includegraphics[width=\columnwidth]{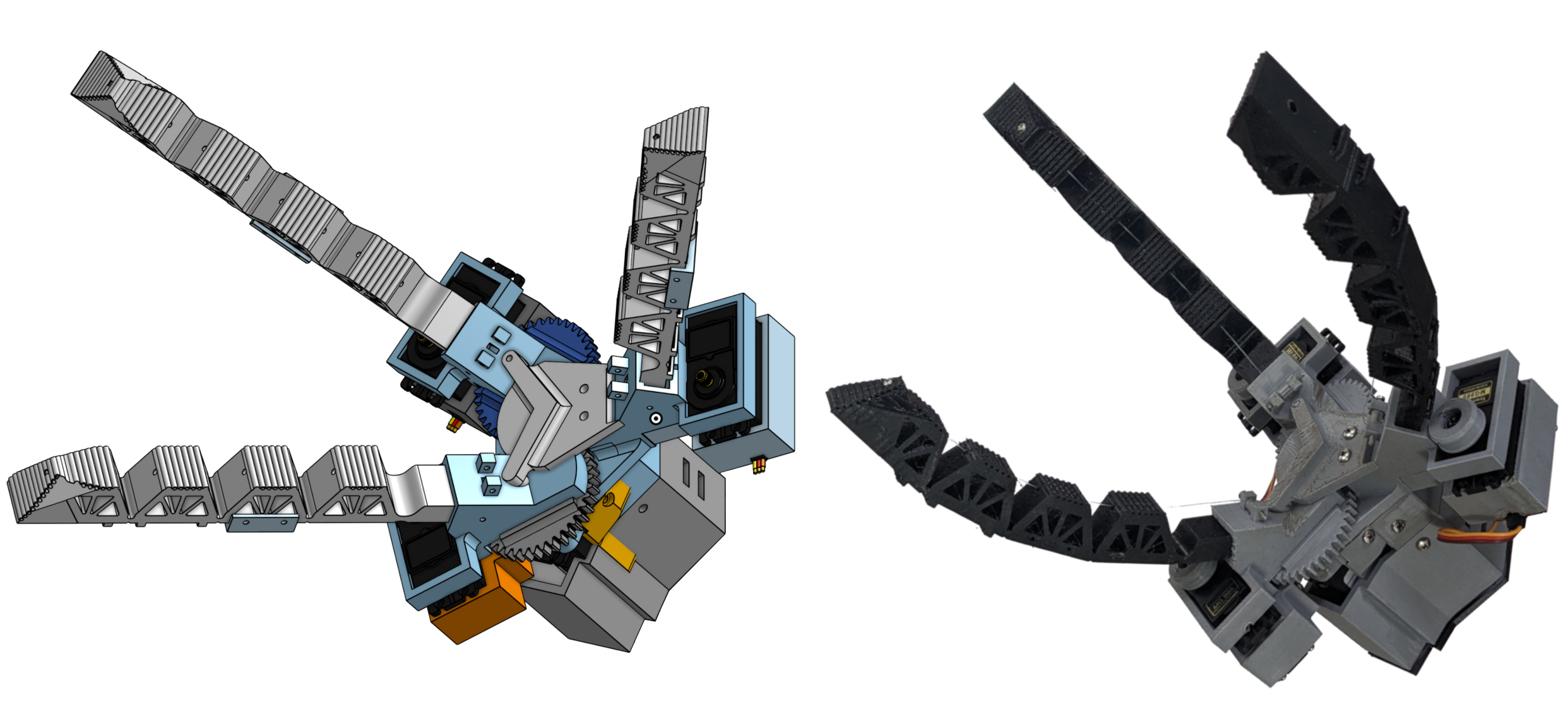}
\caption{CAD model and the 3D printer Gripper}
\end{figure}

\subsubsection{Re-configurable Soft Robotic Gripper}
While the two-fingered grippers are suitable for grasping simple cuboidal rigid items, it is not capable of grasping a wide range of non-cuboidal and non-rigid items. In order to cater to these items, a custom designed three-fingered soft gripper with an ability to reconfigure itself is used. Reconfiguration allows the gripper to shift between cylindrical and spherical grasps, enabling it to grasp a larger spectrum of objects that it may encounter in a retail store setting. Additionally, because of the soft nature of the fingers, the gripper is able to grasp even delicate or compliant objects like farm produce and chip packets which are not suitable for the rigid two-fingered gripper. 

The proposed gripper was designed and fabricated using a 3D printer. It consists of a rigid palm and three soft fingers attached to the palm. The palm and the soft fingers are printed using Polylactic Acid (PLA) and Thermoplastic Polyurethane (TPU-95A) respectively. Each finger is separately actuated using tendons which are further driven by servo motors (one for each finger). The base of one finger among the three is fixed, while the base of the other two can move relative to the palm, the motion of the movable fingers is coupled by means of a meshing gears and is actuated by a single servo motor. Details on the control techniques can be found in our previous works in ~\cite{9926580,10442683}. The gripper is designed in a modular fashion, allowing the soft fingers to be replaced when they are worn out. The fingers could also be replaced by any tendon actuated alternative design, enabling the gripper to be customizable for any variety of objects. 

\subsection{Perception}
The perception part of the system consists of an RGB-D camera, shelf-detection module, object detection and object localization module. We have used an Intel D415 RealSense Depth Camera for capturing the RGB-D video of the scene. Shelf detection is done using aruco markers placed along the boundary of the shelf racks. The object localization module uses the output from the object detection module (Yolo V6 \cite{kapoor2023concept}) and the depth information from the RGB-D image to determine the location of the anomalous object in the 3D robot coordinate frame by transforming it from camera frame to robots base coordinate frame. This 3D coordinate is used by the shared autonomy algorithm as the target locations where operator might want to drive the robot.

\subsection{Base Platform Navigation}
The mobile robot base comprises of an omni-directional platform equipped with four mecanum wheels. This four-wheeled design enables omni-directional movement without requiring a conventional steering system, thereby allowing the robotic system to swiftly move in any direction from any configuration. This mechanisms, also known as holonomic drive, grants the robot the capability to travel in every direction irrespective of its orientation\cite{taheri2015kinematic}. Eq. (\ref{eq:omni_kinematics}) represents the mathematical model of inverse kinematics of the mecanum drive mobile base to obtain angular velocities of the wheels ($\omega_i$) from the demanded body velocities: linear ($v_x$, $v_y$) and angular $\omega_z$.

\begin{figure}[h]
    \centering
    \includegraphics[width=0.6\columnwidth]{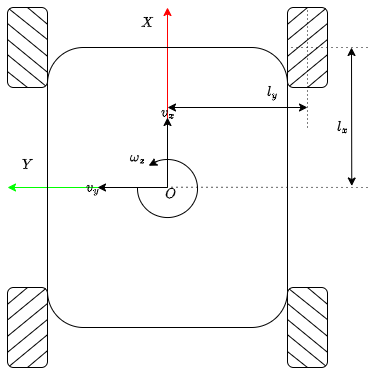}
    \caption{Schematic Diagram of 4-Mecanum Wheel Omni-directional Drive}
    \label{fig:omni_drive}
\end{figure}

\begin{equation}
    \label{eq:omni_kinematics}
    \begin{bmatrix}\omega_1\\\omega_2\\\omega_3\\\omega_4\end{bmatrix} = \frac{1}{R_w}
    \begin{bmatrix}
        1 & -1 & -(l_x+l_y)\\
        1 & 1 & (l_x+l_y)\\
        1 & 1 & -(l_x+l_y)\\
        1 & -1 & (l_x+l_y)
    \end{bmatrix}
    \begin{bmatrix}v_x\\v_y\\\omega_z\end{bmatrix}
\end{equation}

The mobile base is teleoperated using the buttons on the VR controllers. The touchpad of the left hand controller of HTC Vive VR Headset is mapped to the linear xy-motion ($v_x$, $v_y$) of the mobile base and the touchpad of the right hand controller is used to command the yaw ($\omega_z$).

Since the omni-directional platform can move in any direction without needing to align with its heading, it introduces a new challenge in teleoperated control. The motion direction of the mobile base may not align with the robot's heading, potentially limiting the operator's ability to fully perceive the environment for safe navigation. To mitigate this challenge, additional safety measures have been implemented. Utilizing LiDAR data, the system detects obstacles during movement, restricting velocity in the detected direction and halting the robot to prevent accidental collisions.



\section{TELE-OPERATION AND SHARED CONTROL}
\label{sec:teleopeAndSA}
To aid the operator in executing desired actions through teleoperation, such as accessing a particular item on a shelf or manipulating a large object beyond the capability of a single arm, a constrained optimal control problem is devised. This problem is designed to track the operator's commanded inputs while simultaneously assisting in reaching the target position. Additionally, it ensures compliance with collision avoidance and kinematic constraints.

\subsubsection{Cost function\:}
Cost function to achieve the teleoperation with shared control capability is defied as follows
 \begin{align}
 \label{eq:Cost}
\begin{split}
	&J(\mathbf{x},\mathbf{u}; \mathbf{g}) =\\ &\sum_{i}\Big[(\mathbf{x_i}-\mathbf{x_{d_i}})^T\mathbf{K}_{min}\mathbf{Q}(\mathbf{x_i}-\mathbf{x_{d_i}})+ (\mathbf{u_i}-\mathbf{u_{d_i}})^T\mathbf{R}(\mathbf{u_i}-\mathbf{u_{d_i}}) \\&+ \sum_{j}^{|\mathbb{G}|}(\mathbf{^gR_b}(\mathbf{x_i}-\mathbf{g_c}_j))^T(\mathbf{I}-\mathbf{K}(\mathbf{g}(j),\mathbf{x}))\mathbf{P}\left(\mathbf{^gR_b}(\mathbf{x_i}-\mathbf{g_c}_j)\right)\Big]
\end{split}
\end{align}
where, $\mathbf{x}=[\mathbf{x}_L^T,\mathbf{x}_R^T]^T\in R^{6\times 1}$ is the vector formed by concatenation of left and right end-effector positions $(\mathbf{x}_L,\mathbf{x}_R)$ respectively, similarly. $\mathbf{u}=[\mathbf{u}_L^T,\mathbf{u}_R^T]^T\in R^{6\times 1}$is the vector formed by the concatenation of left and right end-effector velocities respectively, $\mathbf{x}_d=[\mathbf{x}_{L_d}^T,\mathbf{x}_{R_d}^T]^T\in R^{6\times 1}$ and $\mathbf{u}_d=[\mathbf{u}_{L_d}^T, \mathbf{u}_{R_d}^T]^T\in R^{6\times 1}$ are the concatenated versions of desired positions and velocities of the end-effectors respectively. $\mathbf{Q}\in R^{6\times 6}$ and  $\mathbf{R}\in R^{6\times 6}$ are respectively positive semi-definite and positive definite weighing matrices. $\mathbf{P}\in R^{6\times 6}$ is a diagonal positive definite weighing matrix. The reference trajectories $(\mathbf{x_d}_L,\mathbf{x_d}_R)$ for both the end-effectors in this case are generated by forward propagation of a constant velocity model $\mathbf{x_d} = {\mathbf{x_{d_i}}: x_{d_i}=x_{d_{i-1}}+\mathbf{v}*\delta}$, wherein, the velocity estimate $\mathbf{v}$ is obtained by using a Kalman filter with a constant velocity model and operator's captured positions as measurements. $\mathbf{K}\in R^{6\times 6}$ is an adaptive weighing matrix which is dependent on the end-effector positions $\mathbf{x}\in R^{6\times 1}$ and goal positions stacked with duplication as $\mathbf{g_c}=[\mathbf{g}^T,\mathbf{g}^T]^T\in R^{6\times 1}$  where $\mathbf{g}\in \mathbb{G}$ with $\mathbb{G}$ being the set of all detected goals.  

The first part of the cost function is given in eq. \ref{eq:Cost} is the reference tracking part which ensures that the motion (position as well as velocity) of the end-effectors are synchronised with that of the operator. The second part of the cost function is goal-tracking, which aims to assist the user in driving the robot accurately to the goal location. This term is designed in such a way that, when any end-effector approaches the goal the diagonal terms of the adaptive weighing term $\mathbf{K}(\mathbf{g_j},\mathbf{x}))$ will converge towards zeros thus minimizing the significance of reference tracking and increasing the weight for goal tracking. Additionally, in order to ensure that the robot approaches the goal only in the desired direction, we first transform the relative position between the end-effector and the goal from the robot base frame to the goal frame using the block diagonal rotation matrix $\mathbf{^gR_b}=blkDiag[\mathbf{^gR_{b_L}},\mathbf{^gR_{b_R}}]\in R^{6\times 6}$ where, $(\mathbf{^gR_{b_L}},\mathbf{^gR_{b_R}})$ are the rotation matrices corresponding to orientation of the goals with respect to the base of the robot followed by scaling it with the diagonal matrix $\mathbf{P}$ to minimise the error in a desired direction more rapidly than the others.

The adaptive scaling matrix $\mathbf{K}(\mathbf{g}(j),\mathbf{x}))$ is a block diagonal matrix computed as follows
\begin{align}
    \mathbf{K}(\mathbf{g_j},\mathbf{x}))&=\begin{bmatrix}
        w(\mathbf{g_j}[:3],\mathbf{x}[:3])\mathbf{I}_3&\mathbf{0}_3\\
        \mathbf{0}_3&w(\mathbf{g_j}[3:],\mathbf{x}[3:])\mathbf{I}_3\\
    \end{bmatrix}\\
    w(\mathbf{t},\mathbf{p}) &= \dfrac{1}{1+\exp(\beta-\alpha||(\mathbf{t}-\mathbf{p})||)}
\end{align}
where, $\mathbf{I}_3\in R^{3\times 3}$ is an identity matrix, $(\cdot)[:3]$ is used to indicate the slicing operation in which elements of the vector up-to specified integer are extracted. $w(\mathbf{g},\mathbf{x})$ is a scalar function that accepts the 3D position of the end effector $\mathbf{p}$ and target goal $\mathbf{t}$ and is designed to smoothly switch from tracking operators commanded trajectory to assisting the user in reaching a goal of interest based on user states. $\mathbf{K}_{min}$ is the block diagonal matrix with its block diagonal matrices corresponding to the minimum block diagonal matrices computed across all the goals, $\alpha$ and $\beta$ are tuning parameters.
\begin{figure}[t]
    \centering
    \includegraphics[width=\columnwidth]{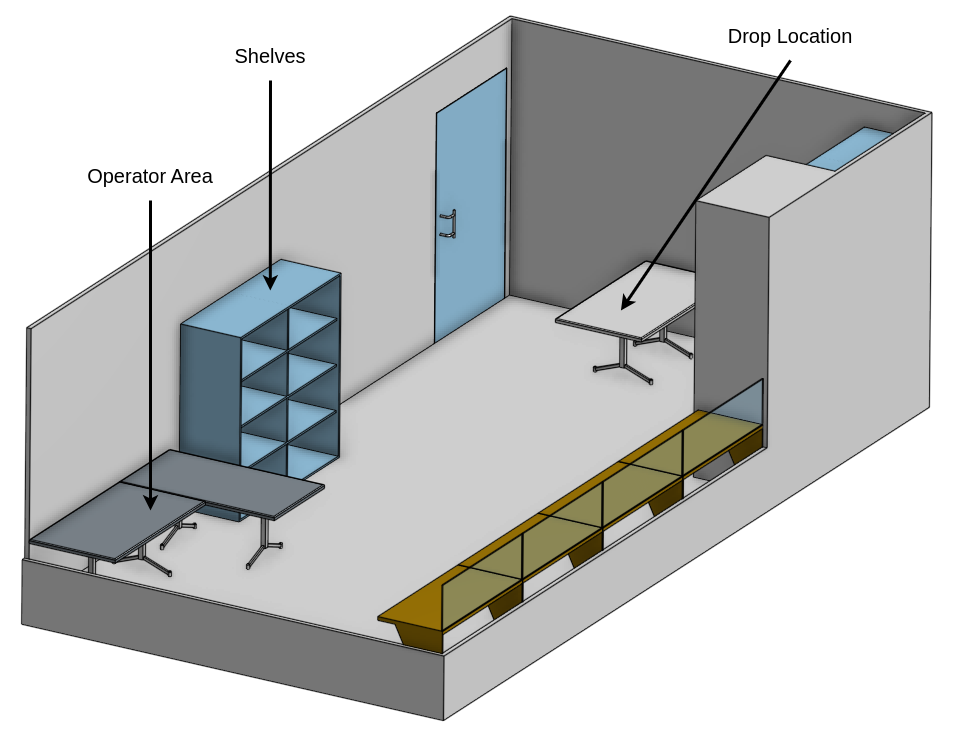}
    \caption{CAD model of mock retail store}
    \label{fig:mockupRetilStore}
\end{figure}

\begin{figure}[t]
\centering
\begin{tabular}{cccc}
\includegraphics[width=0.22\textwidth]{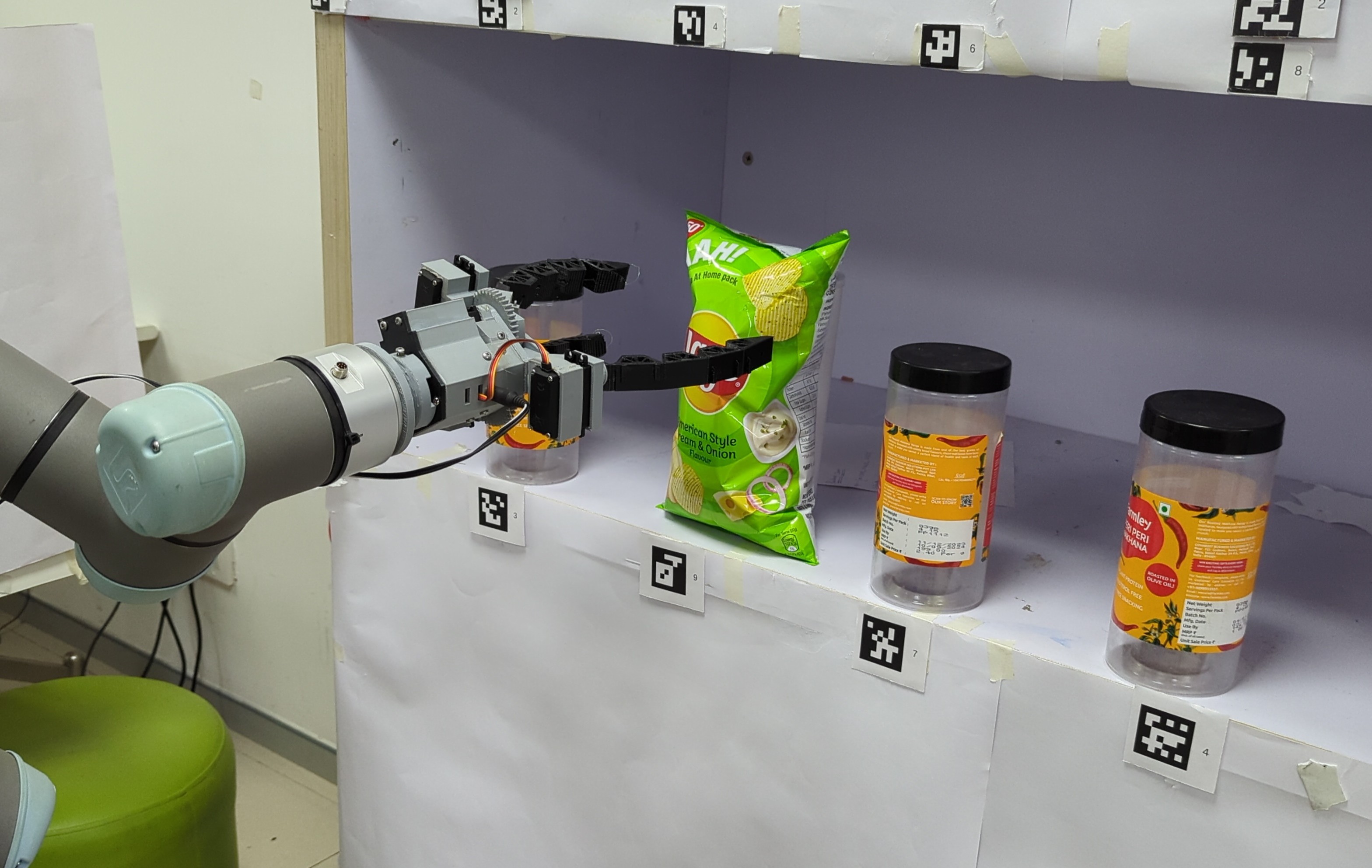} &
\includegraphics[width=0.225\textwidth]{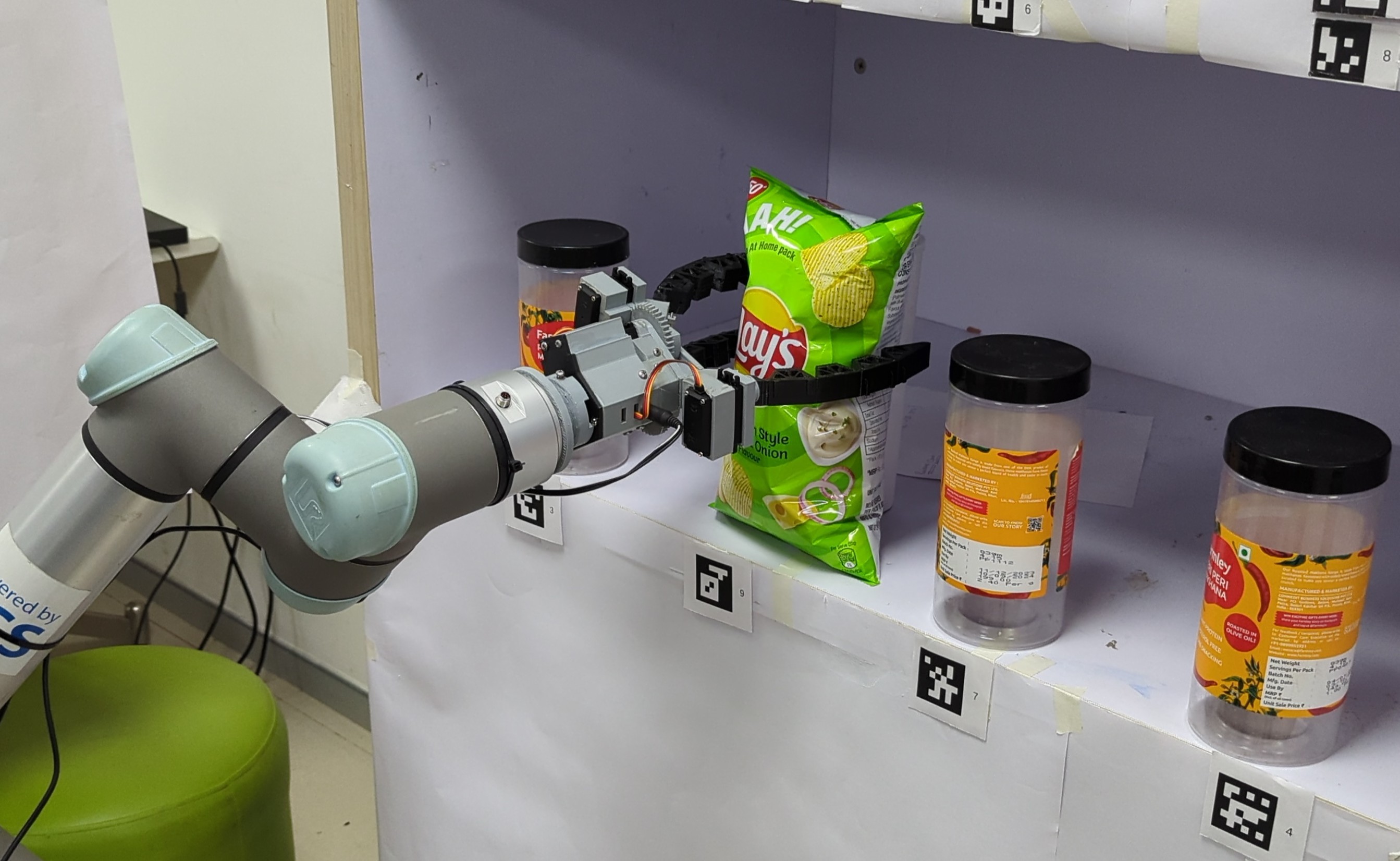} \\
\textbf{(a)}  & \textbf{(b)} \\[6pt]
\end{tabular}
\begin{tabular}{cccc}
\includegraphics[width=0.22\textwidth]{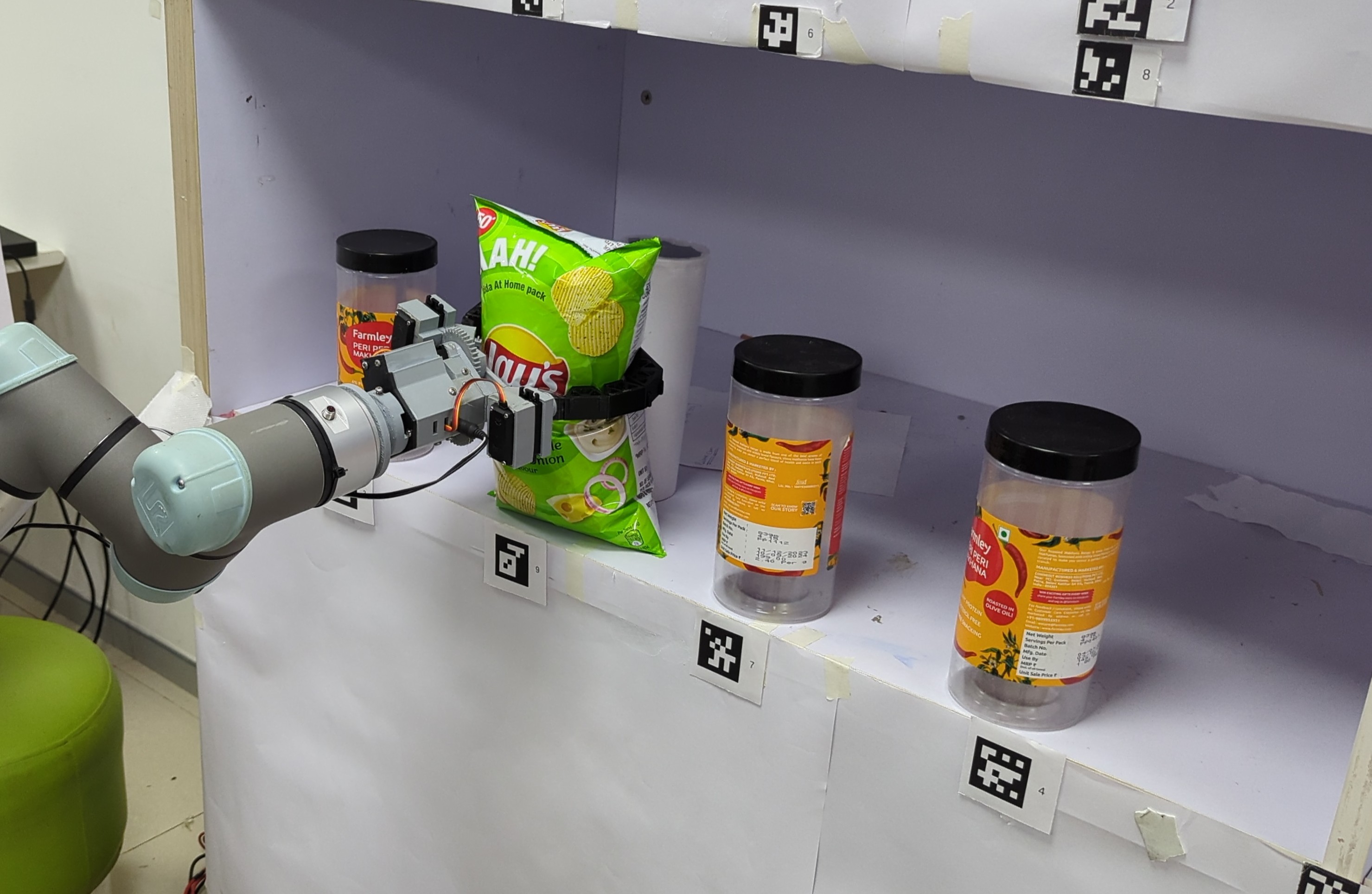} &
\includegraphics[width=0.225\textwidth]{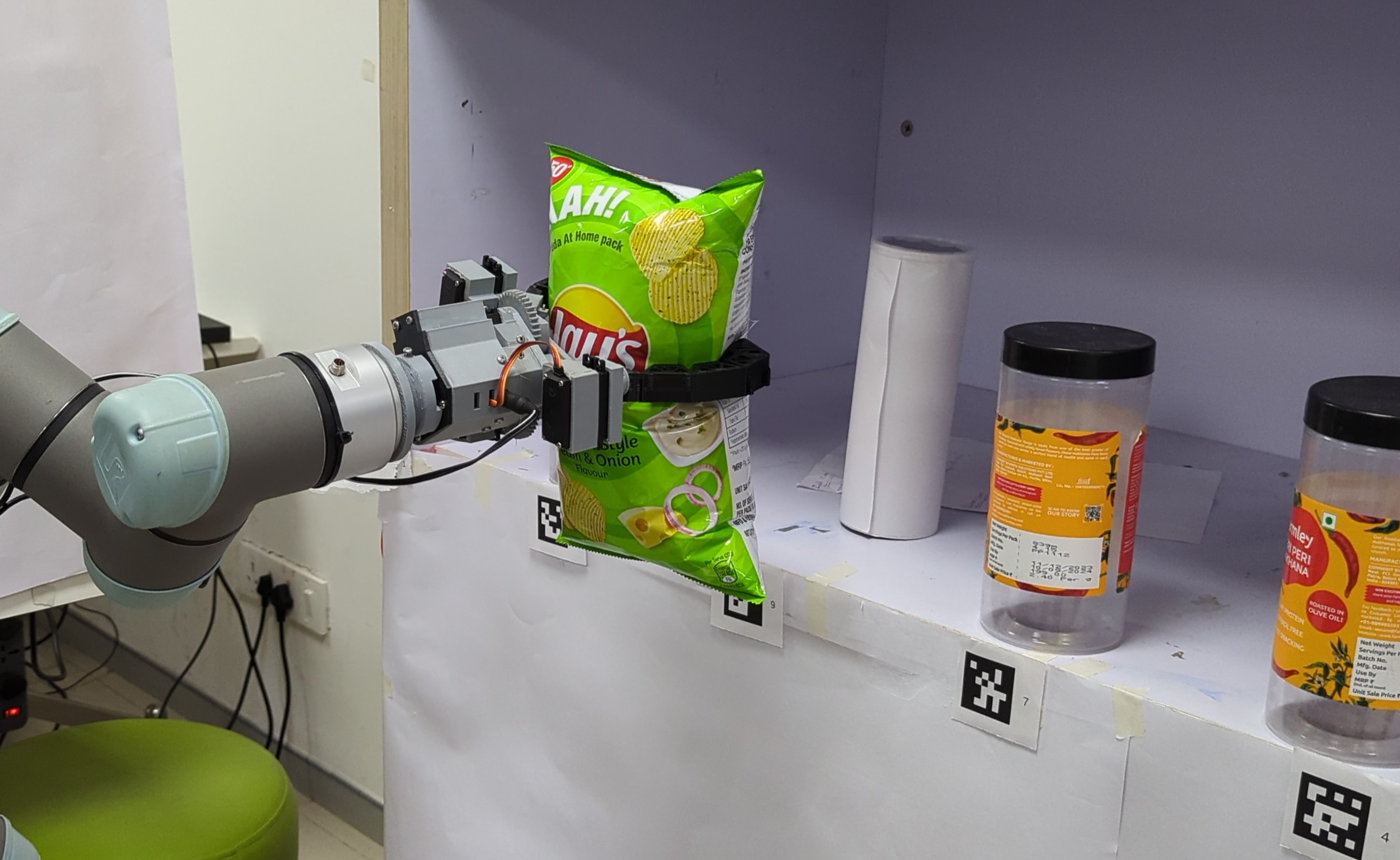} \\
\textbf{(c)}  & \textbf{(d)} \\[6pt]
\end{tabular}
\caption{Snapshots of soft gripper approaching the item in shelf: \textbf{(a)} Start position
\textbf{(b)} Approach position
\textbf{(c)} Gripping Position
\textbf{(d)} Lifting position}
\label{fig:softGrippergrasping}
\end{figure}

\textit{System Model}: A point mass model is used to model the motion of each end-effector independently, this is not a hard requirement and even a double integrator can be used since the trajectory obtained by solving the resulting optimal control problem using MPC methodology is used as a reference for the inner-loop high gain controller, which renders the inner loop system as identity system to outer-loop controller.
\begin{align}
    {\mathbf{x}}_{t+1} = {\mathbf{x}}_t+{\mathbf{u}}_t\delta
\end{align}
where, $\delta = T/N$, with $T$ being the time horizon and $N$ being the number of steps in the horizon.\\
\textit{Constraints}: The end effector's position is constrained to prevent collisions with both the robot body and items on the shelf. This is achieved through collision avoidance constraints, implemented using planners and ellipsoidal constraints. Similarly, constraints to ensure compliance with kinematic requirements at the grasping locations, particularly when the robot engages in coordinated bi-manual manipulation to handle longer items are defined.

A planner constraint is defined for preventing collision with objects such as the end of the shelf, or floor and is defined as
\begin{align}
	\mathbf{A}\mathbf{x}\geq	\mathbf{b}
\end{align}
where, rows of matrix $\mathbf{A}$ comprise the unit normal to the plane and the elements of $\mathbf{b}$ correspond to the calibrated limits. For example, to prevent the end-effector from colliding with the floor, one of the rows of the $\mathbf{A}$ is chosen as $[0,0,1]$ and the corresponding value of $\mathbf{b}$ is chosen as the height from the base of the robot to the floor.

Similarly, to avoid collision with smaller objects, an ellipsoidal constraint is used to restrict the end effector's trajectory from entering the ellipsoidal volume enclosing the objects in the workspace.
\begin{align}
\label{eq:ellipsoid}
(\textbf{x}-\textbf{x}_{o})^T
    \textbf{R}\mathbf{M}\textbf{R}^T(\textbf{x}-\textbf{x}_{o})>\alpha
\end{align}
where $\textbf{x}_o$ corresponds to the centroid of the obstacle to which ellipsoid is fitted and $\mathbf{R}$ is the rotation matrix defining the orientation of the ellipsoid with respect to the base frame of the robot.  $\mathbf{M}$ is a diagonal scaling matrix and $\alpha$ is a scalar value, chosen empirically to enclose the obstacle with-in the ellipsoid.

Further, for enabling the dual arm coordinated manipulation, we define kinematic constraints for the grasping approaches corresponding to the top-down and front-direction approach, the following constraints are used,
\begin{align}
\label{eq:kinemaicConstraint1}
S(||\textbf{x}_L-\textbf{x}_{L_d}||)&=0\\
S(||\textbf{x}_R-(\textbf{x}_L+\textbf{R}_L^L\textbf{x}_R)||)&=0\\
S(\textbf{R}_L^T\textbf{R}_R-\mathbf{I})&=0\\
S(\textbf{R}_L^T\textbf{R}_{L_o}-\mathbf{I})&=0
\end{align}
Where, $S$ is the selection variable set to one, when the goals that the end effectors are approaching are defined on the same object otherwise set to zero,  $\textbf{R}_L$ represents the rotation matrix for representing the orientation of the left end-effector with respect to the base, $^L\textbf{x}_R$ represents the position of the right end-effector in the left frame, $\textbf{R}_{L_o}$ represents operator's commanded position captured using the VR controllers orientation.
\begin{figure}[t]
    \centering
    \includegraphics[width=\columnwidth]{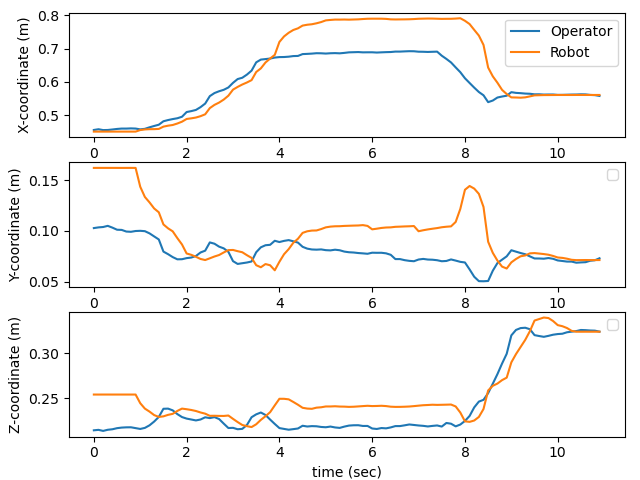}
    \caption{Operators and robots trajectory while grasping soft items}
    \label{fig:leftArmTraj}
\end{figure}
Similarly, for the side grasping approach, the kinematic constraints are defined as 
\begin{align}
\label{eq:kinemaicConstraint2}
S(||\textbf{x}_{L_d}-\textbf{x}_L||)&=0\\
S(||\textbf{x}_R-(\textbf{x}_L+\textbf{R}_L^L\textbf{x}_R)||)&=0\\
S(\textbf{R}_L^T\textbf{R}_{L_o}-\textbf{I})&=0\\
S(\textbf{R}_R^T\textbf{R}_L-\textbf{R}_z(\pi))&=0
\end{align}
where, $\textbf{R}_z(\pi)$ is used as a rotation matrix corresponding to $180\deg$ rotation about z-axis. In both the modes of approach the left end-effector is taken as a reference and the right end-effector is constrained with respect to the left-hand end-effector. 

In addition to this, the velocity of motion of the end-effector is also limited within the fixed range using the bounding constraint as below
\begin{align}
	\mathbf{u}_{\mathbf{min}}\leq	\mathbf{u}\leq	\mathbf{u}_{\mathbf{max}}
\end{align}
Above optimal control problem is solved using a Model Predictive Control methodology using AL-iLQR algorithm ~\cite{howell2019altro}.   
In solving this optimal control problem we get collision-free trajectories for both arms within the specified time horizon, from these trajectories, only the first point is extracted and passed to the inverse kinematic solver (TracIK \cite{beeson2015trac}), which provides  8 joint angles solutions for each arm, from these it is required to select a solution that is closest to the current robots joint configuration. It is also required for the robot to attain a desirable posture/configuration such that the elbow joint is always pointed away from robot's body. Thus to achieve these goals, we use weighted least squares to select the joint positions from the set of solutions obtained from the IK solver.
\begin{align}
\mathbf{\theta}=arg \min_\theta \sum_i^N\ ||\mathbf{\theta}-\mathbf{\theta}_l||^2_{\mathbf{W}_l}+||\mathbf{\theta}-\mathbf{\theta}_d||^2_{\mathbf{W}_d}
\end{align}
where, $||\cdot||_{\mathbf{W}_l}$ represents weighted norm of difference in the IK solution $\mathbf{\theta}$,  $\mathbf{\theta}_l$ and $\mathbf{\theta}_d$ represent the current robots joint position and desired joint configuration respectively with weight matrix $\mathbf{W}_l$ and $\mathbf{W}_d$ respective errors.

\begin{figure}[t]
\centering
\begin{tabular}{cccc}
\includegraphics[width=0.2\textwidth]{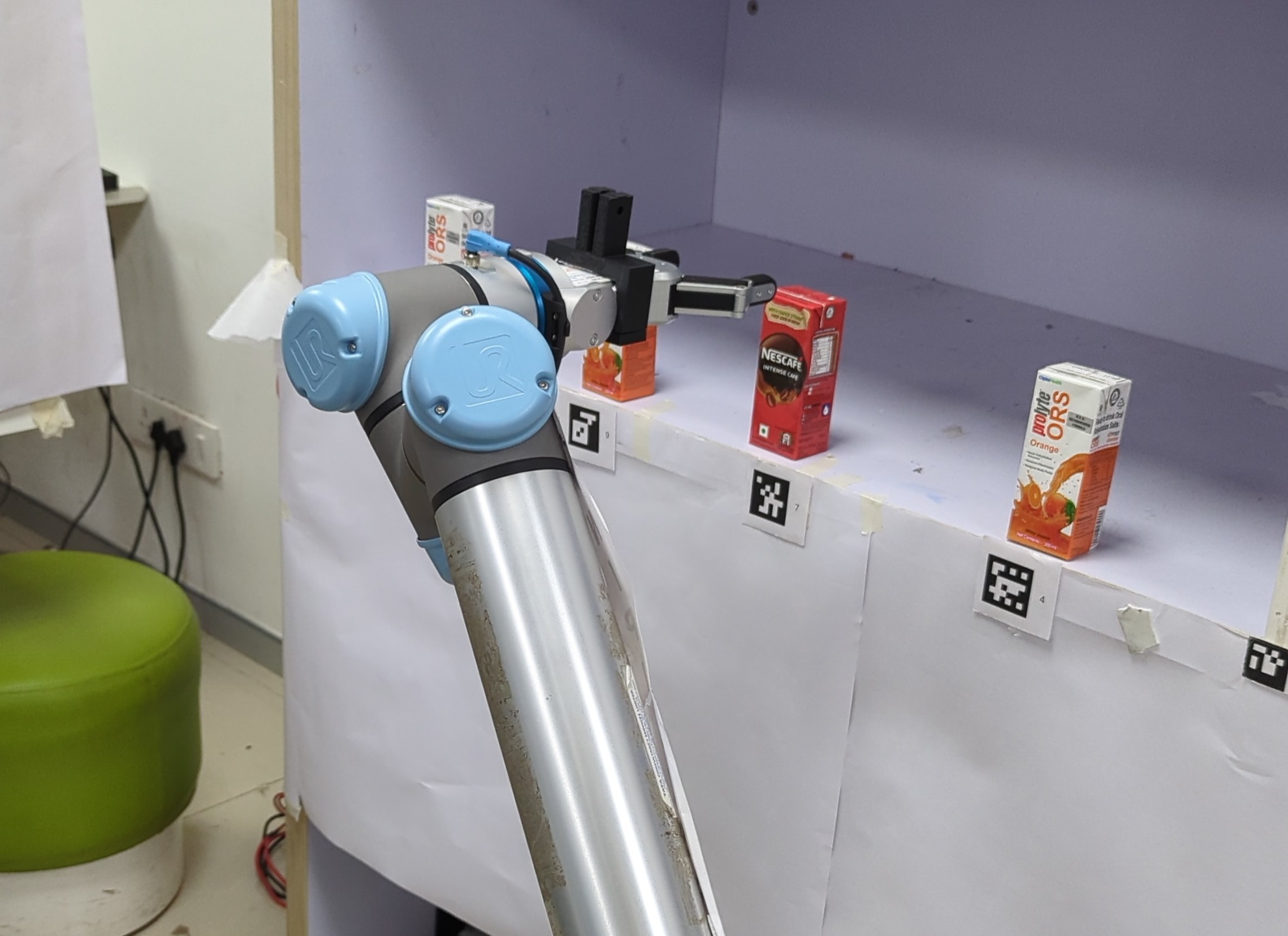} &
\includegraphics[width=0.2\textwidth,trim={0cm 4cm 0 3cm},clip]{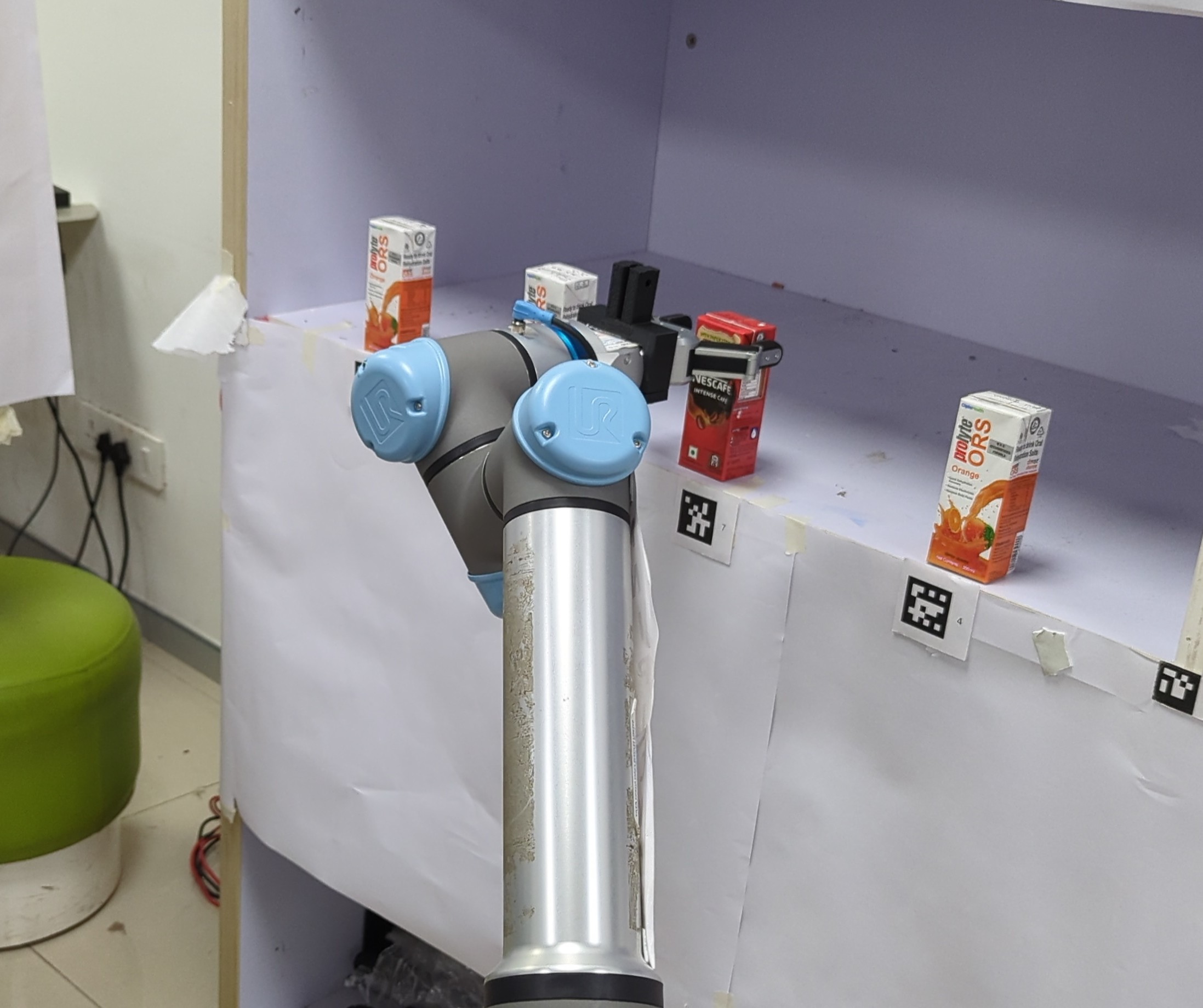} \\
\textbf{(a)}  & \textbf{(b)} \\[6pt]
\end{tabular}
\begin{tabular}{cccc}
\includegraphics[width=0.2\textwidth]{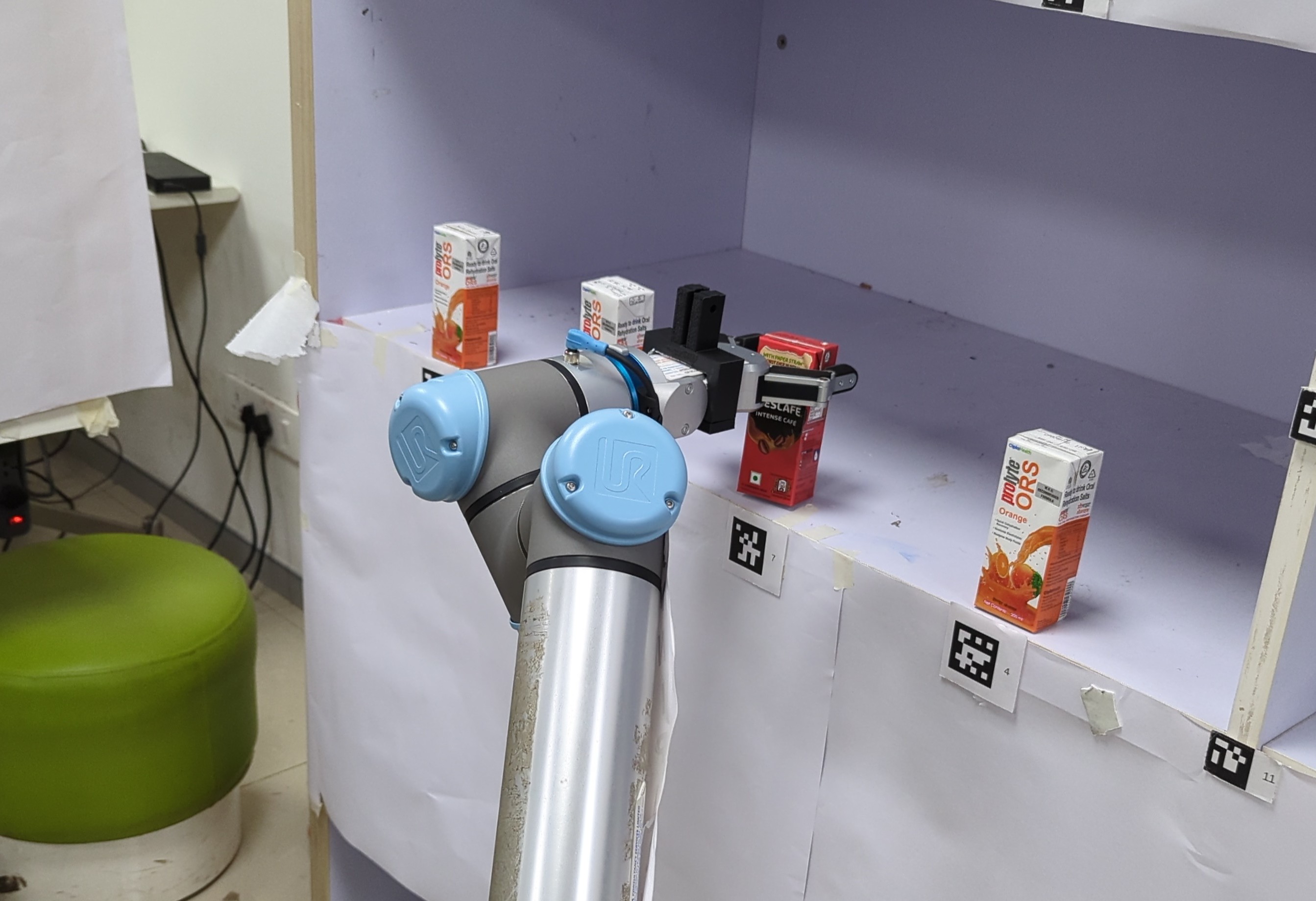} &
\includegraphics[width=0.2\textwidth, trim={0cm 4cm 0 3cm},clip]{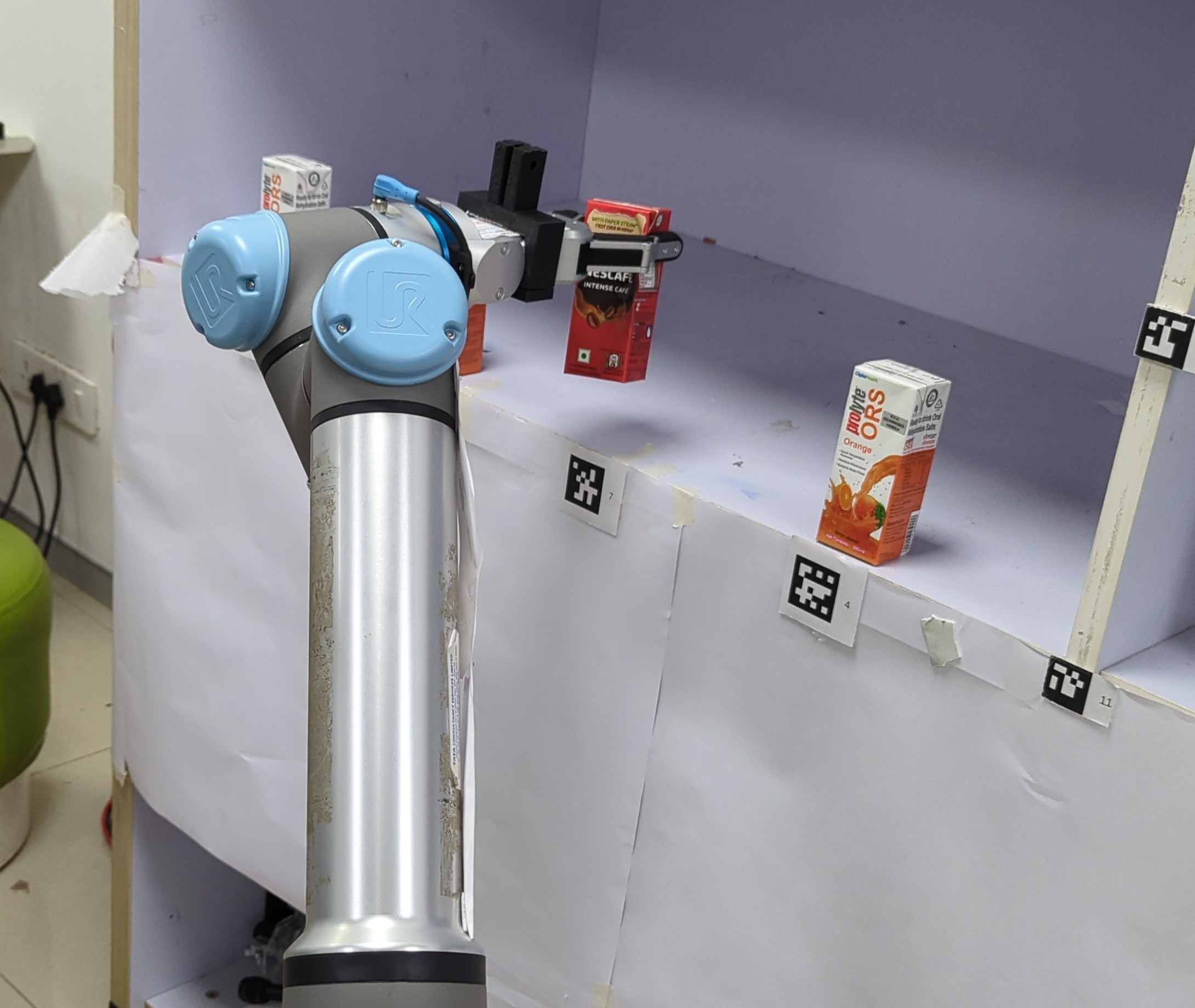} \\
\textbf{(c)}  & \textbf{(d)} \\[6pt]
\end{tabular}
\caption{Snapshots of RG2 gripper approaching the item in shelf: \textbf{(a)} Start position \textbf{(b)} Approach position \textbf{(c)} Gripping Position \textbf{(d)} Lifting position}
\label{fig:RG2Grasping}
\end{figure}
\begin{figure}[t]
    \centering
    \includegraphics[width=0.9\columnwidth]{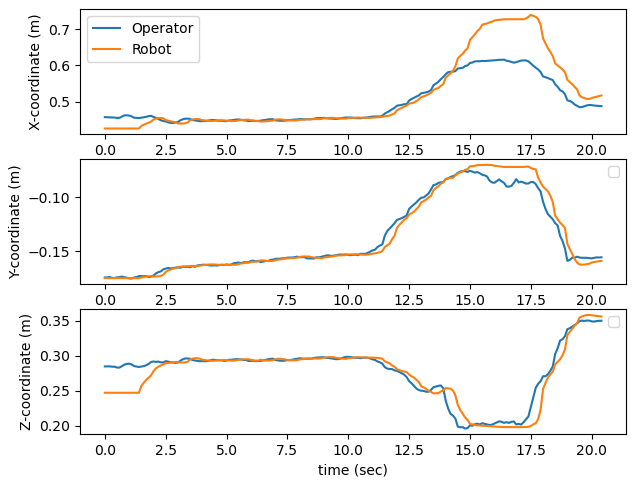}
    \caption{Operators and robots trajectory while grasping soft items}
    \label{fig:rightArmTraj}
\end{figure}
\section{SYSTEM EVALUATION}
\label{sec:sysEval}
In order to evaluate the performance of the proposed teleoperation system, several experiments involving picking items from the shelves in a mock-up retail setting were performed. The details of the setup used and the scenarios used for evaluating the system are discussed below.
\subsection{Retail Scenario}
Evaluation of the robot was performed in a mockup retail store as shows by the CAD model in Fig. \ref{fig:mockupRetilStore}. It comprises of the operator area, a shelf with common retail items and a drop location to drop off any anomalous items picked from the shelf. 

\subsection{Single Arm Manipulation}
Results corresponding to the manipulation of smaller items that require use of a single arm are presented here. A sample snapshots of a soft gripper approaching the item and grasping it is shown in Fig. \ref{fig:softGrippergrasping} and the associated robots and operator trajectory is shown in Fig.\ref{fig:leftArmTraj}. It can be seen from the plot that the robot closely follows the operator's commanded trajectory until it reaches near the item location, which is located approximately at [0.79,0.1,0.25]$m$ with respect to the base frame of the dual-arm robot system. On reaching near the target, the robot switches to autonomous assistance and moves to the item location thus assisting the user in the terminal task that otherwise is time consuming due to limited perception of the remote environment through the VR, this is evident from the plot where the robot's trajectory has deviated from the operator's trajectory in the time range of 4 to 8 $secs$. We can also see that the initial error between the robot and the operators position is minimised by the proposed MPC controller.

Similarly, Snapshots of the second arm fitted with commercially available Onrobots RG2 griper is shown in Fig. \ref{fig:RG2Grasping}, and the associated trajectory is shown in Fig. \ref{fig:rightArmTraj}. The Trajectory plots indicate a similar behaviour to that of soft gripper trajectories.


\begin{figure}[t]
\centering
\begin{subfigure}[b]{0.23\textwidth}
    \centering
    \includegraphics[width=0.8\columnwidth]{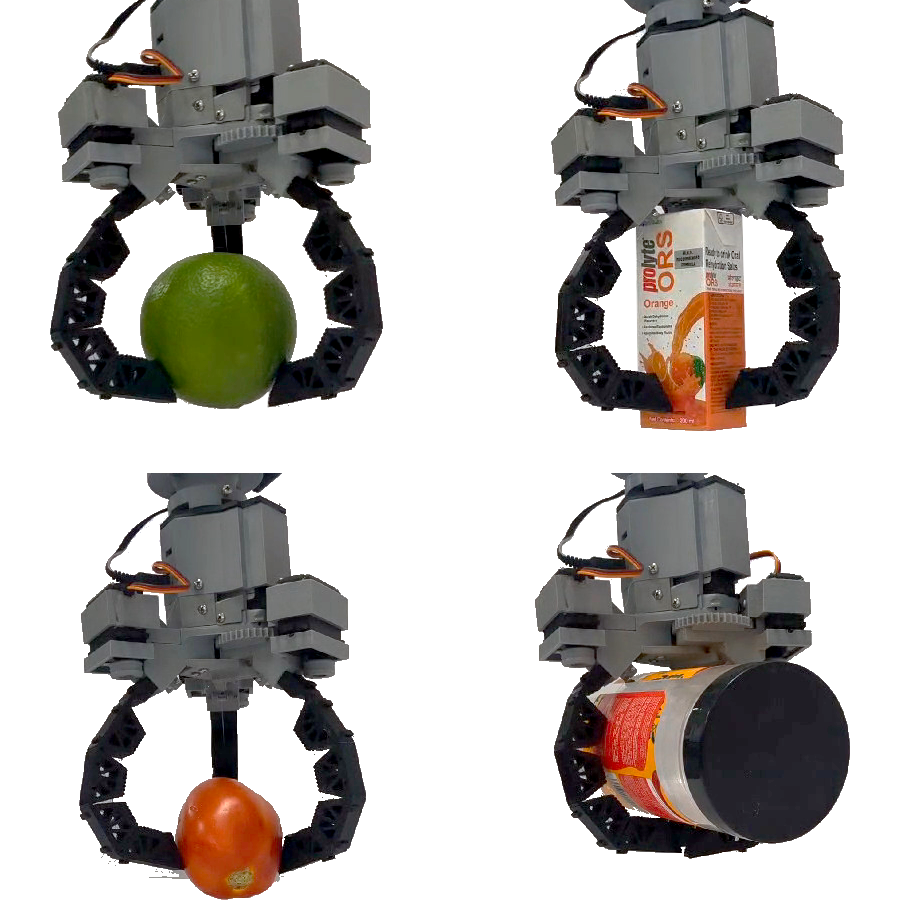}
    \caption{Top-Down grasp of retail objects with spherical (left-top, right-top, left-bottom) and cylindrical(right-bottom) configuration}
    \label{fig:topdown_grasp_fig}
\end{subfigure}
\hfill
\begin{subfigure}[b]{0.23\textwidth}
    \centering
    \includegraphics[width=0.9\columnwidth]{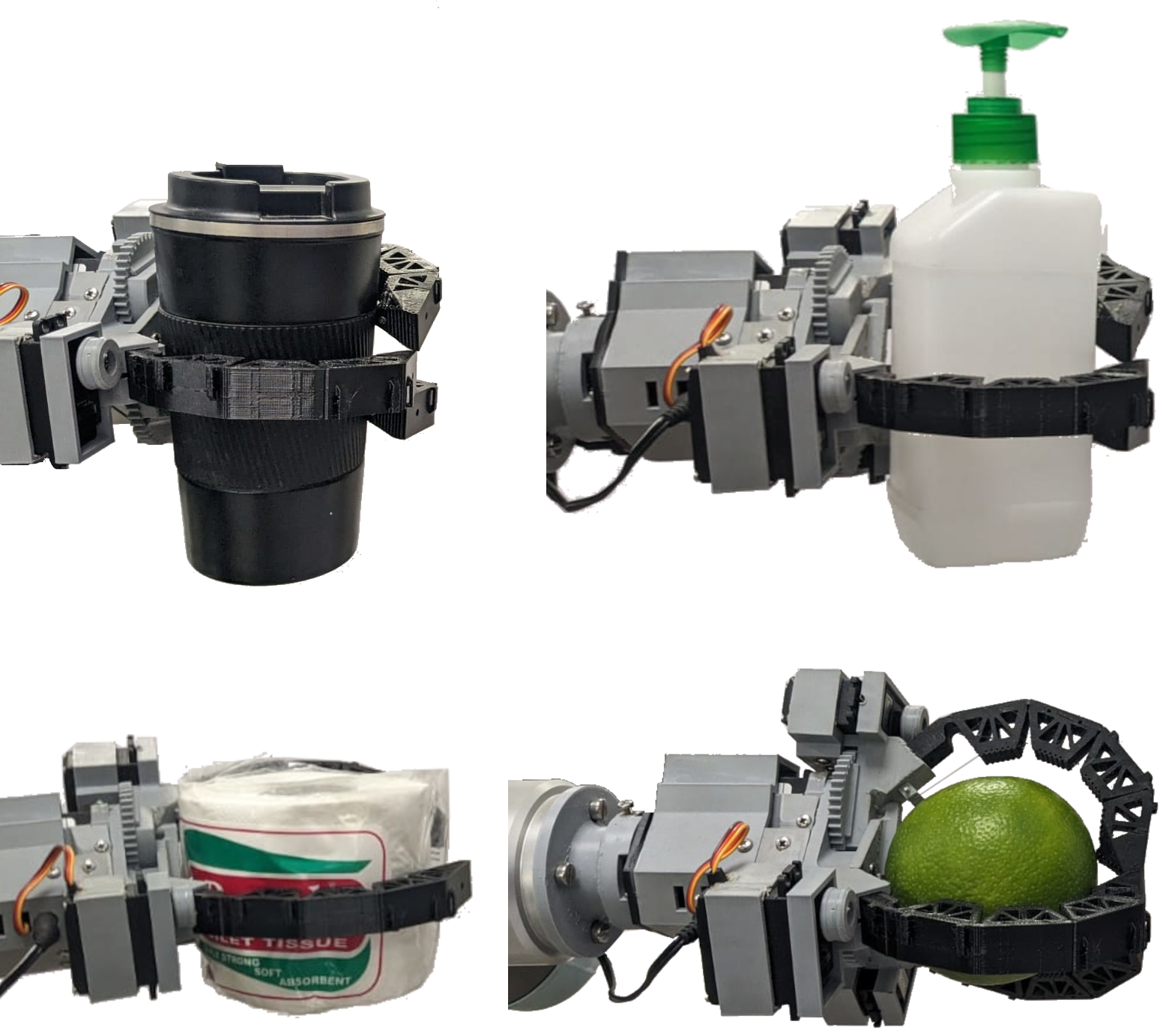}
    \caption{Sideways grasp of retail objects in cylindrical (left-top, right-top, left-bottom) and spherical (right-bottom) configuration }
    \label{fig:sideways_grasp_fig}
\end{subfigure}
\end{figure}

In addition to the teleoperation capability, the custom-designed soft gripper was also extensively tested for its efficacy to grasp retail items. Objects of different shapes and sizes were used to imitate the variety of objects normally encountered in a retail setting, including fruits, vegetables, plastic bottles, tin cans, cuboidal tetra-packs and chips packets. The soft gripper was able to grasp all of the items when used in a top-down approach provided the dimension of the object along the grasp closure direction of the fingers was less than the width of the gripper in open position (18 cm). The compliant nature of the fingers allowed the gripper to conform itself to the shape of the object being grasped and it either led to a pinch grasp or a form closure grasp depending on the size of the object. Sideways grasps proved to be much more challenging because of the inability of the gripper to attain form closure when the objects are too small for the dimension of the soft-fingers. We estimated that for an object with a dimension less than 7 cm along the grasp closure direction, the gripper is not able to attain a successful grasp. However, this limitation can be overcome by scaling the design of the soft-fingers accordingly. Fig. \ref{fig:topdown_grasp_fig} and Fig. \ref{fig:sideways_grasp_fig} depict the successful grasps of common retail items from top-down and sideways approaches respectively. 

\begin{figure}[t]
\centering
\begin{tabular}{cccc}
\includegraphics[width=0.2\textwidth]{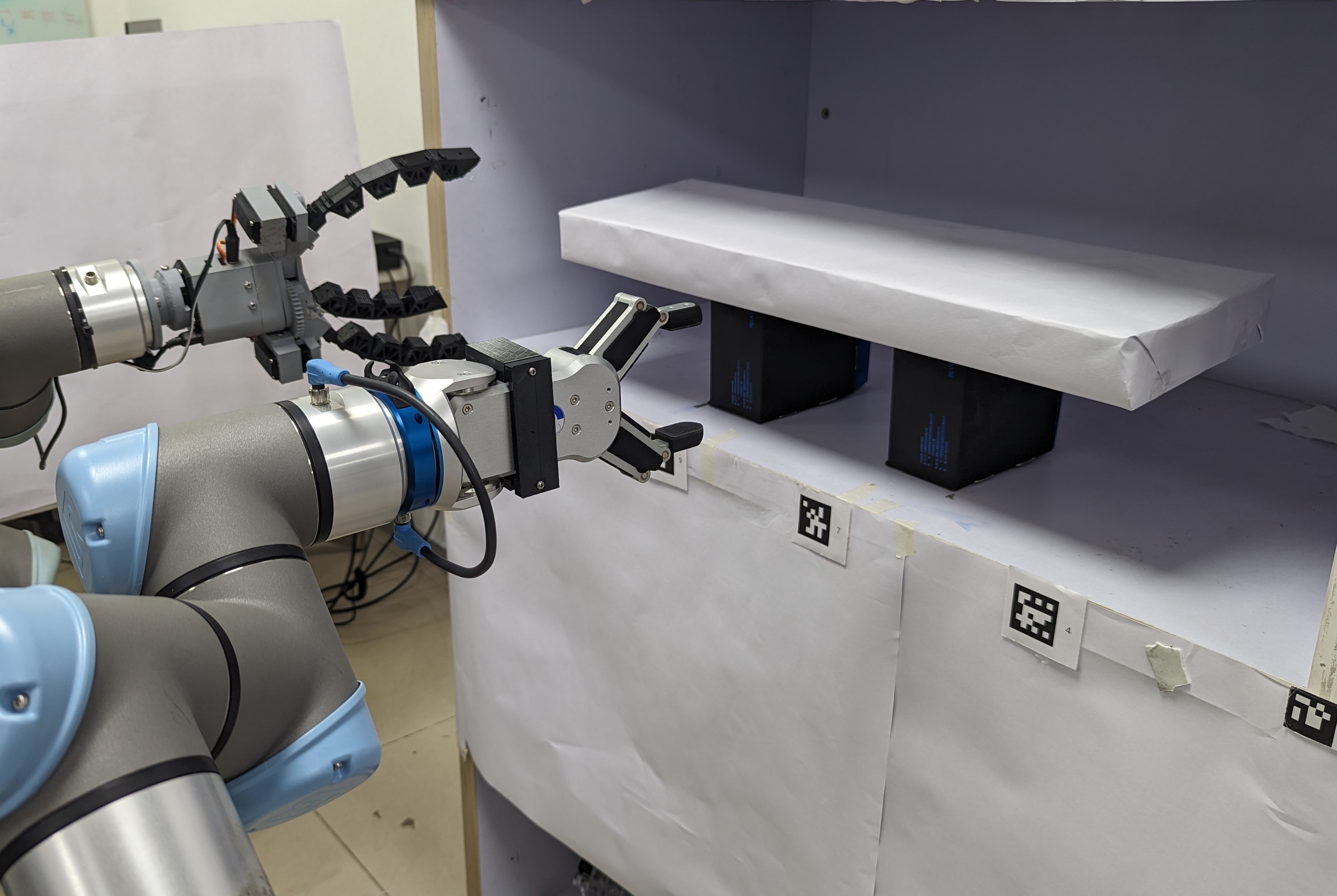} &
\includegraphics[width=0.2\textwidth]{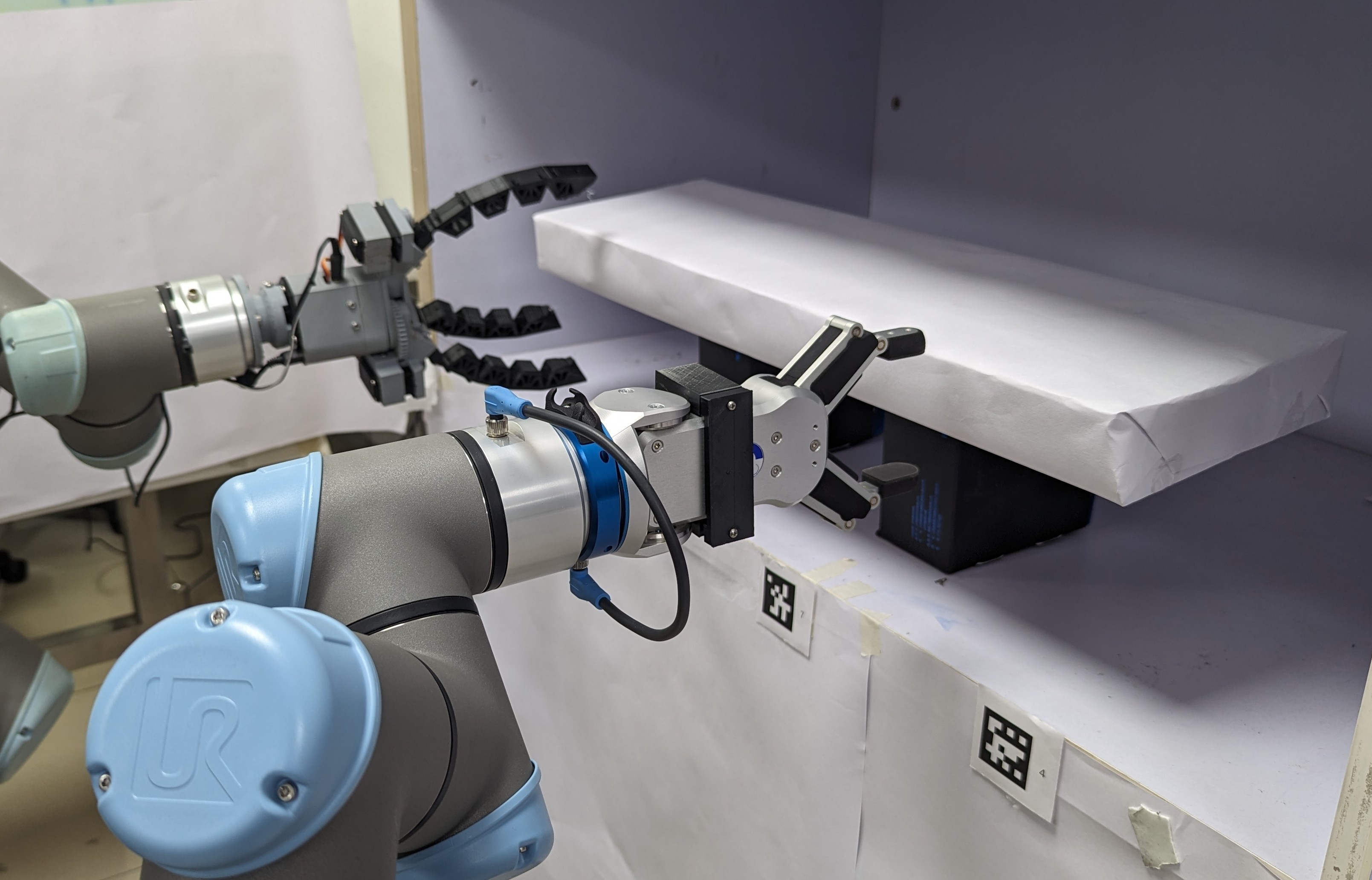} \\
\textbf{(a)}  & \textbf{(b)} \\[6pt]
\end{tabular}
\begin{tabular}{cccc}
\includegraphics[width=0.2\textwidth,trim={0cm 0cm 0cm 0cm},clip]{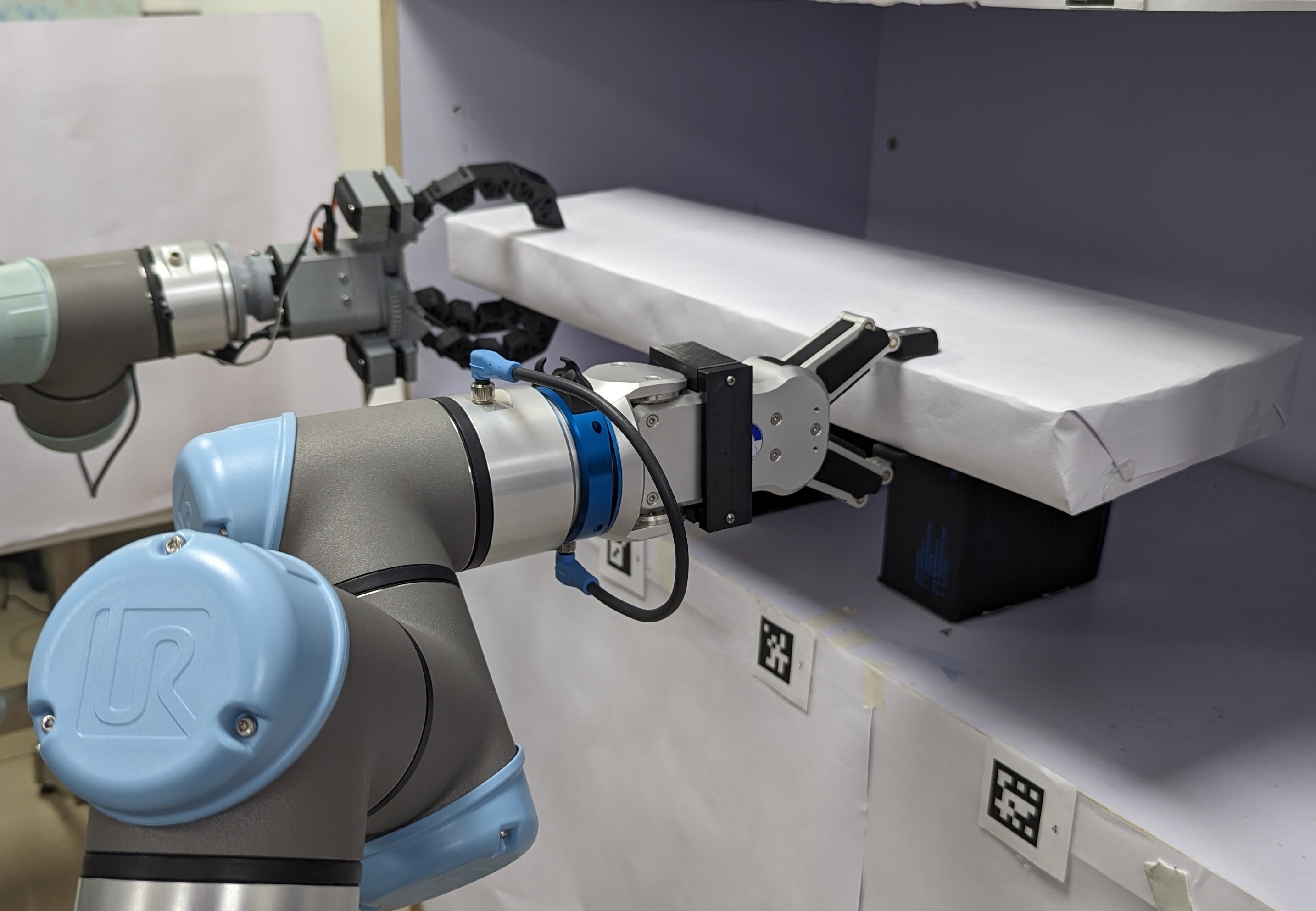} &
\includegraphics[width=0.2\textwidth,trim={5cm 0cm 5cm 0cm},clip]{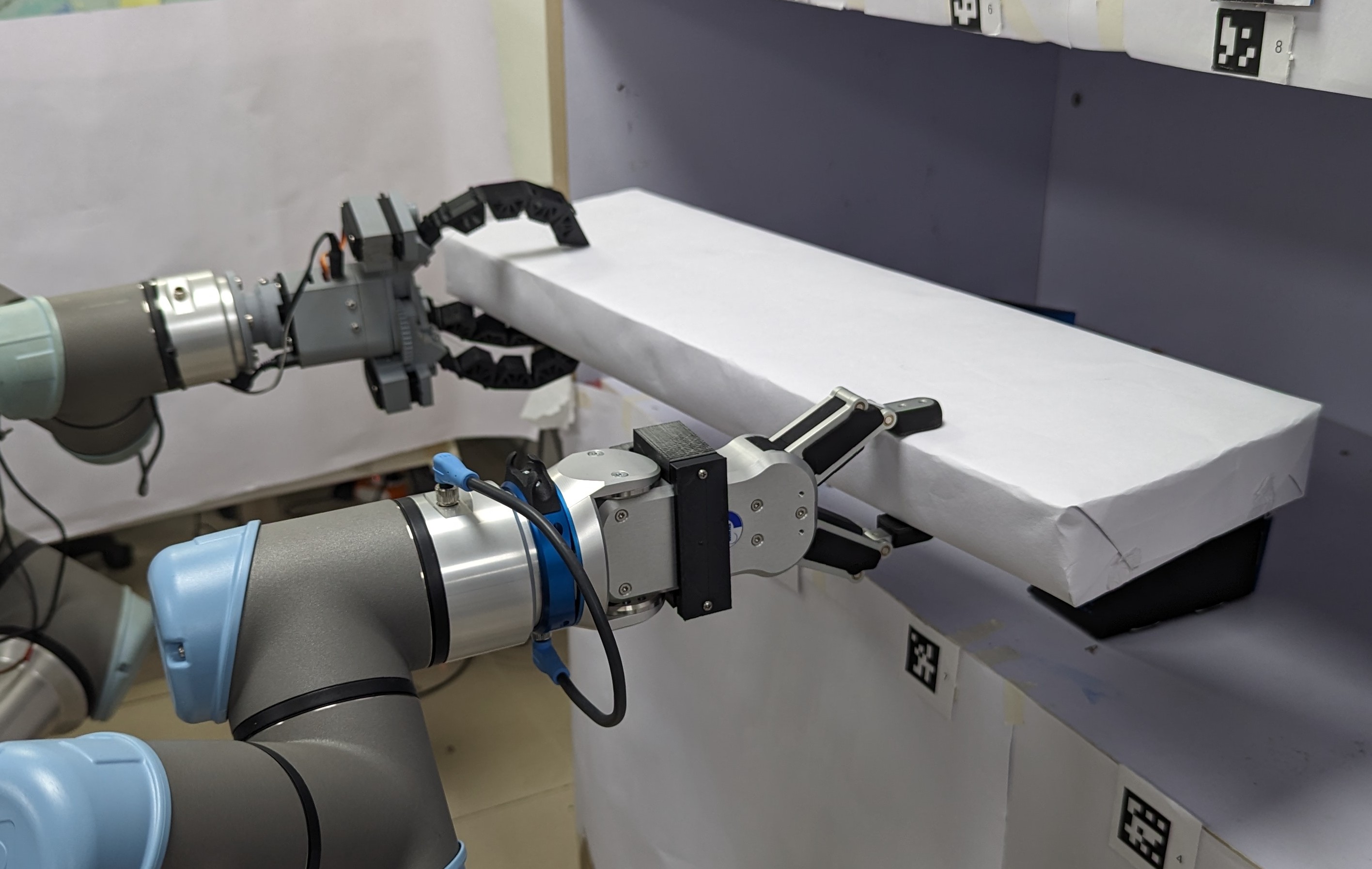} \\
\textbf{(c)}  & \textbf{(d)} \\[6pt]
\end{tabular}
\caption{Snapshots of dual arm operation while grabbing long item in shelf: \textbf{(a)} Start position \textbf{(b)} Approach position \textbf{(c)} Gripping Position \textbf{(d)} Lifting position}
\label{fig:DualArmGrasping}
\end{figure}

\begin{figure}[t]
    \centering
    \includegraphics[width=\columnwidth]{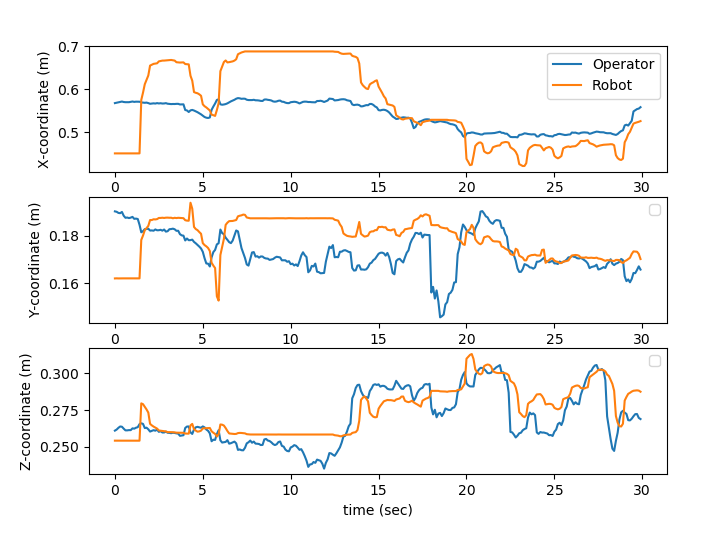}
    \caption{Operator's left commanded and corresponding left end-effector's trajectory while performing dual arm coordinated manipulation}
    \label{fig:dualArmLeftTrj}
\end{figure}

\begin{figure}[t]
    \centering
    \includegraphics[width=\columnwidth]{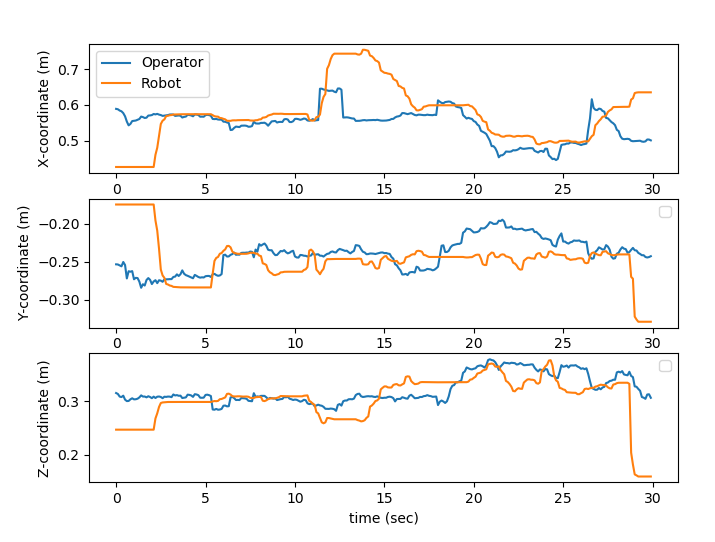}
    \caption{Operator's right commanded and corresponding right end-effector's trajectory while performing dual arm coordinated manipulation}
    \label{fig:dualArmLeftTraj}
\end{figure}

\begin{figure}[t]
    \centering
    \includegraphics[width=0.9\columnwidth, trim={0cm 0.5cm 0.5cm 0.5cm}, clip]{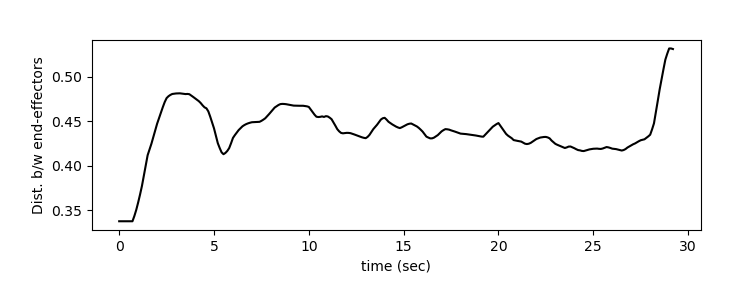}
    \caption{Distance between the two end-effectors while performing dual arm coordinated manipulation}
    \label{fig:DistancePlot}
\end{figure}

\subsection{Dual Arm Manipulation}
Fig. \ref{fig:DualArmGrasping} shows the snapshots of the dual-arm robot performing a picking operation and the associated left and right arm trajectories are given in Fig. \ref{fig:dualArmLeftTrj} and Fig.\ref{fig:dualArmLeftTraj} respectively. It can be seen from the trajectories that the characteristics of both the arms trajectories are similar to that of individual arm trajectories in the single arm cases with the only difference being that the robot stops tracking the operator once the goal is grasped. This is because, robots motion is constrained by the kinematic constraints to maintain the distance and orientation between the two end-effectors. The plot in Fig. \ref{fig:DistancePlot} shows the variation of distance between the two end-effectors during the manipulation, it can be seen that the distance between the two manipulators is maintained almost constant when the item is picked and pulled out from the shelf with the total variation being limited within $5cm$, the variation in the distance between is mainly because of the lower loop rate of the MPC controller being limited to 10Hz due to the limited onboard computation. This can further be reduced by using impedence control strategies as demonstrated in ~\cite{s21144653}. Additionally, it should be noted that the compliance of using a soft gripper allows us to continue the manipulation without damaging the item or robot itself.

It was observed through small comparison analyses of shared control strategy against pure teleoperation that the shared control strategy helps in reducing the task completion time by on an average $30\%$, and eliminates the collisions with obstacle compared to pure teleoperation.
\section {CONCLUSION}
\label{sec:conclusion}
In this study, we introduced an omni-directional mobile robot equipped with dual manipulators featuring heterogeneous grippers tailored for inventory management in retail stores via teleoperation. Our primary objective was to address scenarios where fully autonomous robots fail, such as encountering newly introduced items in the store or experiencing failures in perception algorithms. To tackle such challenges, we devised a teleoperation robot with shared control capabilities. This shared control functionality was achieved by designing a cost function to seamlessly transition between pure teleoperation and assistive control based on the operator's movements. We conducted tests in a retail mockup environment, successfully verifying the system's ability to pick up smaller objects with single-arm capabilities and grasp longer items with dual-arm coordination.

Moving forward, our future work aims to expand upon this research by incorporating operator demonstrations to develop a fully autonomous system capable of continual learning from limited operator input. Additionally, we plan to enhance the user interface and visual feedback for operators by integrating haptic feedback and Virtual Reality features. Furthermore, we intend to conduct an extensive user study to evaluate the effectiveness of shared autonomy, considering factors such as task completion time, system usability, and repeatability.
\bibliographystyle{IEEEtran}

\bibliography{teleop_litrature}

\end{document}